# TEXTURE MEASURES COMBINATION FOR IMPROVED MENINGIOMA CLASSIFICATION OF HISTOPATHOLOGICAL IMAGES

**Omar S. Al-Kadi**[*]

*Department of Informatics, University of Sussex, Brighton, BN1 9QJ United Kingdom*

**Abstract**

Providing an improved technique which can assist pathologists in correctly classifying meningioma tumours with a significant accuracy is our main objective. The proposed technique, which is based on optimum texture measure combination, inspects the separability of the RGB colour channels and selects the channel which best segments the cell nuclei of the histopathological images. The morphological gradient was applied to extract the region of interest for each subtype and for elimination of possible noise (e.g. cracks) which might occur during biopsy preparation. Meningioma texture features are extracted by four different texture measures (two model-based and two statistical-based) and then corresponding features are fused together in different combinations after excluding highly correlated features, and a Bayesian classifier was used for meningioma subtype discrimination. The combined Gaussian Markov random field and run-length matrix texture measures outperformed all other combinations in terms of quantitatively characterising the meningioma tissue, achieving an overall classification accuracy of 92.50%, improving from 83.75% which is the best accuracy achieved if the texture measures are used individually.

*Keywords*─ coloured texture analysis, feature extraction, histopathological images, meningioma, naïve Bayesian classifier, Bhattacharyya distance

## 1. Introduction

Meningiomas are one of the most recurring tumours which affect the central nervous system [1-3]. These types of tumours, which have a variable growth potential, develop from the meninges – hence the naming – which are the membranes that cover the brain and spinal cord, and usually do not metastasise (i.e. spread) beyond the location where they originate [4]. It is one of the only brain tumours more common in women than in men – represented by a one to three man to women ratio, and in general, in 94% of the cases the tumour is benign, and in the remaining 2% and 4% it is considered malignant and aggressive; respectively [5].

A means of inspecting histopathological characteristics at a molecular or cellular level is the motivation for the use of microscopic imaging. Although it is an invasive procedure, this modality has the advantage of providing coloured high resolution images exposing the richness or denseness of the examined underlying texture as compared to other non-invasive imaging modalities. This can assist in giving a better interpretation to histopathological images, through studying the effect of disease on the cellular characteristics of the body tissue. This is done by previously staining the extracted tissue biopsies with dyes for visual contrast improvement, which will then facilitate the delineation of cell nuclei, giving a better tissue characterisation. Despite this modality being invasive, which is unpleasant for patients, physicians usually require a biopsy for a definite answer if they are suspicious about a certain abnormality in an image acquired by a non-invasive imaging modality, and a closer view of the histopathological specimens can assist in verifying the tumour type.

Pathologists have been using microscopic images to study tissue biopsies for a long time, relying on their personal experience on giving decisions on the healthiness state of the examined biopsy. This includes distinguishing normal from abnormal (i.e. cancerous) tissue, benign versus malignant tumours and identifying the level of tumour malignancy. Nevertheless, variability in the reported diagnosis may still occur [6-8], which could be due to, but not limited to, the heterogeneous nature of the diseases (i.e. not all samples referring to a certain tumour subtype look identical, raising the issue of misclassification); ambiguity caused by nuclei overlapping; noise arising from the staining process of the tissue samples; intra-observer variability, i.e. pathologist not being able to give the same reading of the same image at more than one occasion; and inter-observer variability, i.e. increase in classification variation between different pathologists. Therefore, through the past three decades, quantitative techniques have been developed for computer-aided diagnosis, which aim to assist pathologists in the process of cancer diagnosis [9]. Currently, the challenge remains in developing a better technique that not just automates the diagnostic procedure, but also applies the optimum texture feature extraction that better captures and understands the underlying physiology to improve cancer recognition accuracy.

A number of research studies have been applied to histopathological images for different tumours in an attempt to automate the diagnosis procedure. Some of them relied on using one texture measure (i.e. method) for feature extraction, such as extraction of wavelet-based features [10-12], or using other measures individually like fractal dimension or grey-level co-occurrence matrix for classification [13, 14]. Using more than one measure for classification was applied as well, such as using spatial and frequency texture features for classification by regression trees analysis [15]. Some used morphological characteristics for feature extraction [16, 17], and others focused more on classifier improvement [18, 19]. Regarding meningiomas, some used unsupervised learning techniques for training artificial neural networks, e.g. a self organizing map, for classifying meningioma features derived by wavelet packet (WP) transform [10]. An average accuracy of 79% was reported for classifying four different meningioma subtypes. Others applied a supervised learning method for classification of meningioma cells [20], using a decision tree after selecting the most relevant features from a base of grey and coloured image features. Also in another two studies the performance of features extracted from four meningioma subtypes using adaptive WP transform was compared to local binary patterns [21] and to co-occurrence methods [22]. The WP method gave the highest classification accuracy of 82.1% when features were classified via a support vector machine classifier after applying a principal component analysis for dimensionality reduction.

As there is very limited research in the literature on fully-automating meningioma classification with significant accuracy, this research sets out to provide a novel method that combines model and statistical-based texture measures in an endeavour to provide a better understanding on how they relate to the underling physiology. The aim is to improve the classification accuracy by integrating the RGB colour channels that better assists the morphological process in segmenting the tumour structure with the best combination of texture features that best captures the characteristics of the examined case.

We intend to seek possible answers to several questions on histopathological image classification: a) will selecting the appropriate colour channel contribute to improving texture classification? b) which texture combinations –after removal of highly correlated features – perform better than the best of the individual measures and why (i.e. how do they relate to underlying physiology)? c) will using multiple texture extraction methods (e.g. more than two methods) guarantee a higher classification accuracy? d) and what is the effect of noise on histopathological images?

The paper is organised as follows. Section II explains the applied technique, and section III will show the experimental results. Then an analysis of the applied texture measures behaviour followed by a discussion is presented in sections IV and V; respectively. Finally, section VI summarises the major outcomes.

## 2. Methodology

Two main stages are involved in the technique used in this paper. In the pre-feature extraction stage, the best colour channel that maximises cell nucleus structure separation from the background is selected, then this colour channel will be used in the morphological processing of all histopathological specimens (i.e. training and testing images). In the next stage, texture features are extracted and fused in all possible combinations, and the optimum features are selected for classification. The complete process is depicted in Fig. 1.

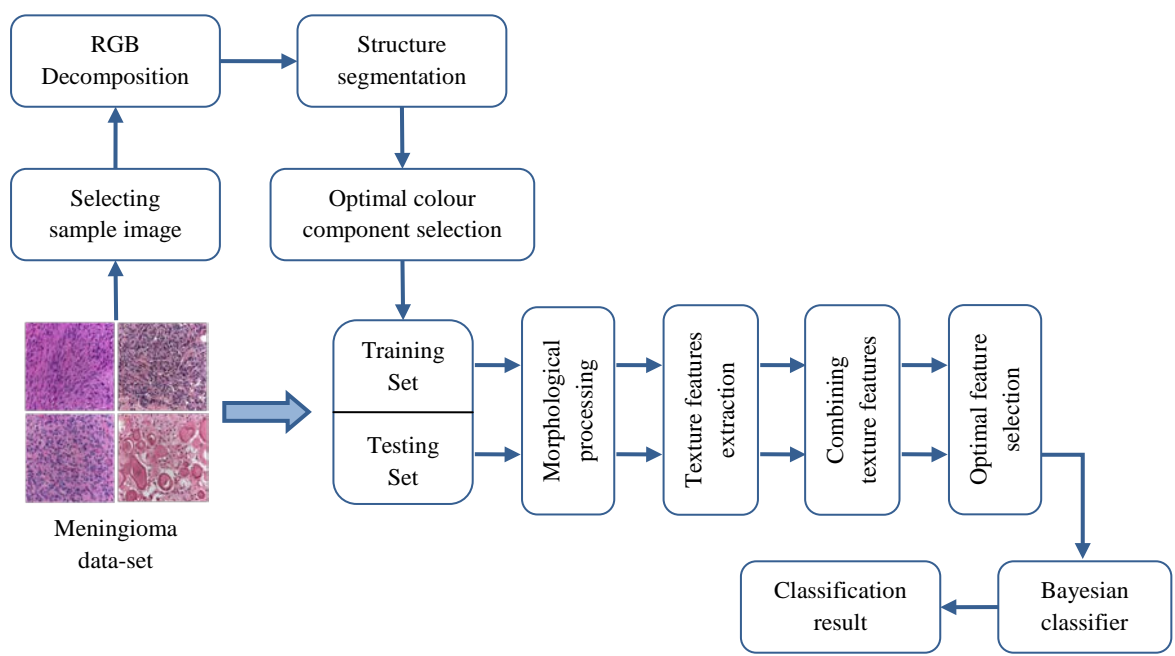

Fig. 1 Diagram explaining the process followed in classification of the histopathological meningioma images.

2.1 Image acquisition and preparation

Four subtypes of grade I meningioma tissue biopsies distinguished according to the World Health Organisation (WHO) grading system [23] are used in this study (Fig. 2). Each subtype has its own features (Table I) which pathologists look for in the processes of tumour classification. The diagnostic tumour samples were derived from neurosurgical resections at the Bethel Department of Neurosurgery, Bielefeld, Germany for therapeutic purposes, routinely processed for formalin fixation and embedded into paraffin. Four micrometer thick microtome sections were dewaxed on glass slides, stained with Mayer's haemalaun and eosin (H&E), dehydrated and cover-slipped with mounting medium (Eukitt®, O. Kindler GmbH, Freiburg, Germany). Archive material of cases from the years 2004 and 2005 were selected to represent typical features of each meningioma subtype. Slides were analysed on a Zeiss Axioskop 2 plus microscope with a Zeiss Achroplan 40×/0.65 lens. After manual focusing and automated background correction, 1300 × 1030 pixels, 24 bit, true colour RGB pictures were taken at standardised 3200 K light temperature in TIF format using Zeiss AxioVision 3.1 software and a Zeiss AxioCam HRc digital colour camera (Carl Zeiss AG, Oberkochen, Germany). Five typical cases were selected for each diagnostic group and four different photomicrographs were taken of each case, resulting in a set of 80 pictures. Each original picture was truncated to $1024 \times 1024$ pixels and then subdivided in a $2 \times 2$ subset of $512 \times 512$ pixel pictures. This resulted in a database of 320 sub-images for further analysis. All acquired images were fully anonymised and our work did not influence the diagnostic process or the patients' treatment.

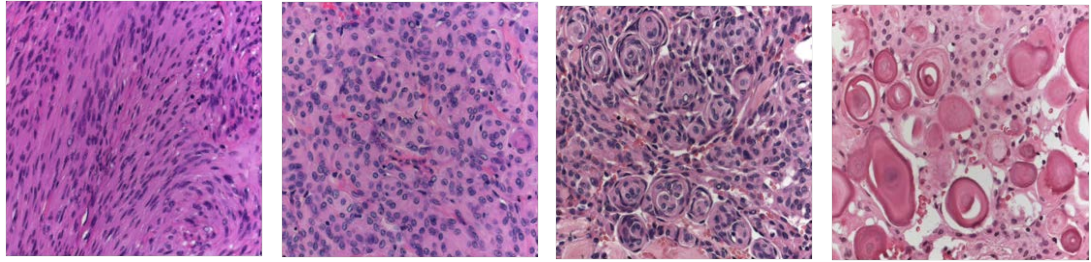

Fig. 2 Four types of grade 1 meningioma, from left to right (fibroblastic, meningothelial, transitional and psammomatous).

Table I Main histological features for the four meningioma subtypes in Fig. 2

| Subtype | Characteristics |
|---|---|
| Fibroblastic | Spindle-shaped cells resembling fibroblasts in appearance, with abundant amounts of pericellular collagen. |
| Meningothelial | Broad sheets or lobules of fairly uniform cells with round or oval nuclei. |
| Transitional | Contains whorls, few psammoma bodies and cells having some fibroblastic features (i.e. spindle-shaped cells) |
| Psammomatous | A variant of transitional meningiomas with abundant psammoma bodies and many cystic spaces. |

2.2 RGB space colour segmentation

Numerous colour spaces have been developed for the purpose of catering for particular applications, e.g. printing, CRT display, television transmission, etc., each of which has its own way for specifying colours for visualisation. The RGB (red, green and blue) colour space is most commonly and widely used in display devices of computer systems and other video related applications for its simplicity and appropriateness for the system's hardware. Other colour spaces such as the HSV (hue, saturation and value) can be more intuitive for human perception – as it coincides more with how we respond to different colour information; nevertheless, this might not be necessarily true for automated classification when we want the computer to take a decision on behalf of us, whereas coloured image segmentation can generally achieve better results using the RGB colour space [24]. Therefore the coloured histopathological images were decomposed to the red, green and blue colour channels to investigate which colour would better distinguish the cell nuclei from the background, hence assisting in improving the classification accuracy in the subsequent morphological and texture feature extraction stages.

One image $I(x, y, z)$, where $z$ is the RGB colour component at point $(x, y)$, is randomly selected from each of the four meningioma subtypes. Then a simple segmentation procedure is applied by first determining the mean RGB vector $a_z = [a_R, a_G, a_B]$ of a sample cell nucleus (see Fig.3), where each component of the vector represents the corresponding colour channel mean. This vector will be used in segmenting each of the colour channel images on a pixel-by-pixel basis. As recommended by [24], the size of the operation box used in the segmentation process was chosen to be 1.25 times the standard deviation $\sigma_z$ of the corresponding colour component of the selected cell nucleus sample values. Shown in (1), each of the pixels were then classified as either 1 or 0 by comparison with $a_z$ giving the segmentation mask which when multiplied with the corresponding colour channel will give the segmented image $I_{seg}$.

Accepted paper in Pattern Recognition 43 (2010) 2043–2053

Also a reference image $I_{ref}$ is generated for each of the randomly selected subtype images after manually segmenting the cell nuclei. Fig. 4 shows $I_{seg}$ and $I_{ref}$ for the blue colour channel of the transitional meningioma subtype.

$$I_{seg}(x,y,z) = \begin{Bmatrix} 1 & if\ I(x,y,z) \in (a_z \pm 1.25\sigma_z) \\ 0 & Otherwise \end{Bmatrix} \quad (1)$$

The Bhattacharyya distance was used to assess the quality of segmentation. For classes with a Gaussian distribution, the Bhattacharyya distance $B_{I_1 I_2}$ is used to estimate the upper bound of classification error between feature image pairs as in (2) [25]. A smaller error value indicates improved separability between the reference and the segmented image. Finally, the colour component with the best segmentation output would be selected for morphological processing.

$$B_{I_1 I_2} = \tfrac{1}{8}(\mu_1-\mu_2)^T \left(\tfrac{\Sigma_1 \Sigma_2}{2}\right)^{-1}(\mu_1-\mu_2) + \tfrac{1}{2} ln\left(\tfrac{|\Sigma_1+\Sigma_2|}{2\sqrt{|\Sigma_1||\Sigma_2|}}\right)$$

$$P_{\varepsilon rror} = \sqrt{P(I_1)P(I_2)}\ exp(-B_{I_1 I_2}) \quad (2)$$

Here, $|\Sigma_i|$ is the determinant of $\Sigma_i$, and $\mu_i$ and $\Sigma_i$ are the mean vector and covariance matrix of class $I_i$ (which refers to $I_{ref}$ and $I_{seg}$).

We will see later that the selection of the colour component of texture plays a significant role in classification accuracy depending on the colour of the examined structure, where other colour spaces will also be examined.

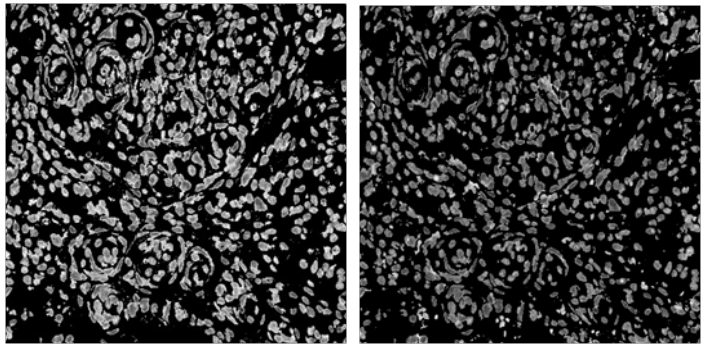

**Fig. 3 Left to right, segmented and reference image for transitional meningioma subtype.**

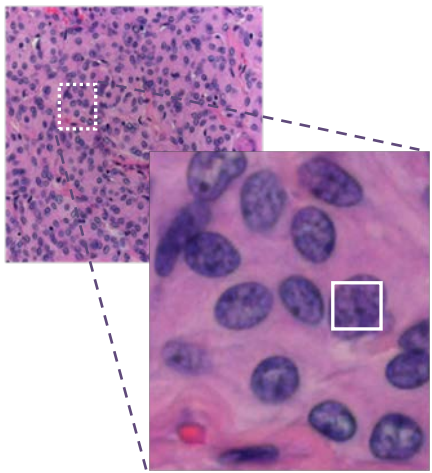

**Fig. 4 The solid white rectangle in the magnified region indicates the selected sample cell nucleus used for segmenting all relevant meningothelial subtypes.**

2.3 Morphological processing

Having selected the appropriate colour channel, morphological processing is required to make the cell nuclei more distinguishable from the background, which has also another advantage which is the elimination of possible noise occurrence [26]. All sets of images are pre-processed by computing the morphological gradient ($M_g$), which is simply the difference between the dilation and erosion of each processed image. It assists in highlighting the edges of the general structure of the texture, which is shown in Fig. 5. To simplify indexing, the image $I(x, y)$ is translated instead of the structuring element $k$ [24].

The grey-scale dilation of image $I(x, y)$ by structuring element $k(x, y)$ can be regarded as the function of all displacements $(s, t)$, such that $I(x, y)$ and $k(x, y)$ overlap by at least one element, this is emphasised in (3) as $(s - x)$ and $(t - y)$, and $x$ and $y$ have to be in domain $I$ and $k$; respectively.

$$(I(x,y) \oplus k(x,y))(s,t) = max\{I(s - x, t - y) + k(x,y) \mid (s - x), (t - y) \in D_I; (x,y) \in D_k\} \quad (3)$$

$D_I$ and $D_k$ are the domains of $I$ and $k$, respectively. The structuring element $k(x, y)$ operates in analogy to a convolution kernel applied to an image. We empirically chose the size of the structure element to be a square 5 x 5 pixels of ones.

Similarly, grey-scale erosion of $k(x, y)$ by image $I(x, y)$ is the function of all displacements $(s, t)$ such that $I(x, y)$, translated by $(s, t)$, is contained in $k(x, y)$.

$$(I(x,y) \ominus k(x,y))(s,t) = min\{I(s + x, t + y) - k(x,y) \mid (s + x), (t + y) \in D_I; (x,y) \in D_k\} \quad (4)$$

Hence $M_g$ can be represented as:

$$M_g = (I(x,y) \oplus k(x,y)) - (I(x,y) \ominus k(x,y)) \quad (5)$$

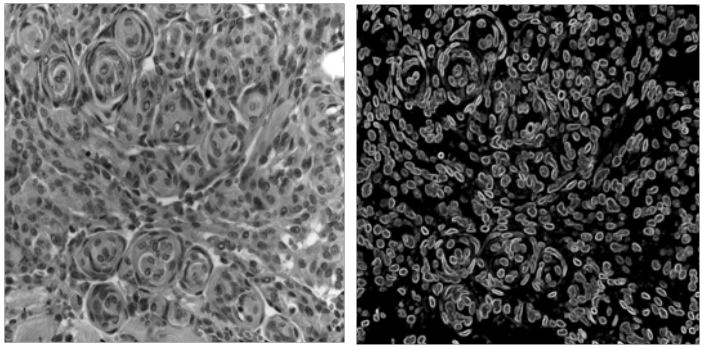

**Fig. 5 Blue colour channel image and its corresponding morphological gradient.**

2.4 Feature extraction approaches

Four different methods – two model and two statistical based – were used to extract different texture features from 320 image samples referring to four meningioma subtypes, as follows:

*2.4.1 Model-based features methods*

*2.4.1.1 Random fields*

Based upon the Markovian property, which is simply the dependence of each pixel in the image on its neighbours only, a Gaussian Markov random field model (GMRF) for third order Markov neighbours was used [27]. Seven GMRF parameters were estimated using the least square error estimation method.

The GMRF model is defined by the following formula:

$$p(I_{xy}|I_{kl}, (k,l) \in N_{xy}) = \frac{1}{\sqrt{2\pi\sigma^2}} exp\left\{-\frac{1}{2\sigma^2}\left(I_{xy} - \sum_{l=1}^{n}\alpha_l s_{xy;i}\right)^2\right\} \qquad (6)$$

The right hand side of (6) represents the probability of a pixel $(x,y)$ having a specific grey value $I_{xy}$ given the values of its neighbours, $n$ is the total number of pixels in the neighbourhood $N_{xy}$ of pixel $I_{xy}$, which influences its value, $\alpha_l$ is the parameter with which a neighbour influences the value of $(x,y)$, and $s_{xy;i}$ is the sum of the values of the two pixels which are in symmetric position about $(x,y)$ and which influence the value of $(x,y)$ with identical parameters (see Fig. 6) where,

$$s_{xy;1} = I_{x-1,y} + I_{x+1,y} \quad s_{xy;3} = I_{x-2,y} + I_{x+2,y} \quad s_{xy;5} = I_{x-1,y-1} + I_{x+1,y+1}$$
$$s_{xy;2} = I_{x,y-1} + I_{x,y+1} \quad s_{xy;4} = I_{x,y-2} + I_{x,y+2} \quad s_{xy;6} = I_{x,y-2} + I_{x+1,y-1}$$

For an image of size $M$ and $N$ the GMRF parameters $\alpha$ and $\sigma$ are estimated using a least square error estimation method, as follows:

$$\begin{pmatrix}\alpha_1\\\vdots\\\alpha_n\end{pmatrix} = \left\{\sum_{xy}\begin{bmatrix}s_{xy;1}s_{xy;1} & \cdots & s_{xy;1}s_{xy;n}\\\vdots & \ddots & \vdots\\s_{xy;n}s_{xy;1} & \cdots & s_{xy;n}s_{xy;n}\end{bmatrix}\right\}^{-1}\sum_{xy}I_{xy}\begin{pmatrix}s_{xy;1}\\\vdots\\s_{xy;n}\end{pmatrix} \qquad (7)$$

$$\sigma^2 = \frac{1}{(M-2)(N-2)}\sum_{xy}\left[I_{xy} - \sum_{l=1}^{n}\alpha_l s_{xy;l}\right]^2 \qquad (8)$$

| $I_{x+2,y-2}$ | $I_{x+2,y-1}$ | $I_{x+2,y}$ | $I_{x+2,y+1}$ | $I_{x+2,y+2}$ |
|---|---|---|---|---|
| $I_{x+1,y-2}$ | $I_{x+1,y-1}$ | $I_{x+1,y}$ | $I_{x+1,y+1}$ | $I_{x+1,y+2}$ |
| $I_{x,y-2}$ | $I_{x,y-1}$ | $I_{xy}$ | $I_{x,y+1}$ | $I_{x,y+2}$ |
| $I_{x-1,y-2}$ | $I_{x-1,y-1}$ | $I_{x-1,y}$ | $I_{x-1,y+1}$ | $I_{x-1,y+2}$ |
| $I_{x-2,y-2}$ | $I_{x-2,y-1}$ | $I_{x-2,y}$ | $I_{x-2,y+1}$ | $I_{x-2,y+2}$ |

**Fig. 6** Third order Markov neighbourhood (in dark) for a sample image pixel $I_{xy}$, compared to second and fourth order Markov neighbourhood represented by the 3 × 3 inner box and the 5 × 5 outer box, respectively.

*2.4.1.2 Fractals*

Fractals are used to describe non-Euclidean structures that show self-similarity at different scales [28]. There are several fractal models used to estimate the fractal dimension (FD); the fractal Brownian motion (fBm) which is the mean absolute difference of pixel pairs as a function of scale as shown in (9) was adopted [29]:

$$E(\Delta I) = K \Delta r^H \qquad (9)$$

where $\Delta I = |I(x_2, y_2) - I(x_1, y_1)|$ is the mean absolute difference of pixel pairs; $\Delta r = \sqrt{(x_2 - x_1)^2 + (y_2 - y_1)^2}$ is the pixel pair distances; $H$ is called the Hurst coefficient; and $K$ is a constant.

The FD can then be estimated by plotting both sides of (9) on a log-log scale and $H$ will represent the slope of the curve that is used to estimate the FD as: *FD = 3 - H*

By operating pixel by pixel, an FD image was generated for each meningioma image where each pixel has its own FD value. Then first order statistical features were derived, which are: mean, variance, lacunarity (i.e., variance divided by mean), skewness and kurtosis.

*2.4.2 Statistical-based feature methods*

*2.4.2.1 Co-occurrence matrices*

The grey level co-occurrence matrix (CM) $P_G(i,j|\delta,\theta)$ represents the joint probability of certain sets of pixels having certain grey-level values. It calculates how many times a pixel with grey-level *i* occurs jointly with another pixel having a grey value *j*. By varying the displacement vector $\delta$ between each pair of pixels many CMs with different directions $\theta$ can be generated. For each sample image segment and with $\delta$ set to one, four CMs having directions (0°, 45°, 90° &135°) were generated.

Having the CM normalised, we can then derive eight second order statistical features which are also known as Haralick features [30] for each sample, which are: contrast (CON), correlation (COR), energy (ENG), entropy (ENT), homogeneity (HOM), dissimilarity (DIS), inverse difference momentum (IDM), maximum probability (MP).

*2.4.2.2 Run-length matrices*

The grey level run-length matrix (RLM) $P_R(i,j|\theta)$ is defined as the number of runs with pixels of grey level *i* and run length *j* for a given direction $\theta$ [31]. RLMs were generated for each sample image segment having directions (0°, 45°, 90° & 135°), then the following eleven statistical features were derived: short run emphasis (SRE), long run emphasis (LRE), grey level non-uniformity (GLN), run length non-uniformity (RLN), run percentage (RP), low grey level runs emphasis (LGLRE), high grey level runs emphasis (HGLRE), short run low grey level emphasis (SRLGLE), short run high grey level emphasis (SRHGLE), long run low grey level emphasis (LRLGLE) and long run high grey level emphasis (LRHGLE).

2.5 Feature selection by correlation thresholding

All extracted features in the combined texture measures were checked for possibly highly correlated features. This process assists in removing any bias towards certain features which might afterwards affect the classification procedure. Although each texture measure tends to characterise the examined texture from a different perspective, some extracted features arise to behave similarly. Another advantage is the alleviation of the curse of dimensionality of texture features [32], which will decrease the computational time and memory required.

An approach which is based on the summation of the divergence measure $D_i$ for each feature $f_i$ between the four different meningioma subtypes was adopted. An advantage of using the divergence function for inspecting feature separability is that it places no prior assumption on class-conditional densities, and has a direct relation with Bayes error [33], which can be defined by the following formula:

$$D_i(f_i) = \sum_{k=1}^{n_c} \sum_{l>k}^{n_c} \frac{(\sigma_{k,f_i} - \sigma_{l,f_i})^2 (1 + \sigma_{k,f_i} + \sigma_{l,f_i})}{2\sigma_{k,f_i}\sigma_{l,f_i}} \qquad (10)$$

where $n_c$ is the number of subtypes – 4 for this work – and $\sigma_{k,f_i}$ and $\sigma_{l,f_i}$ are the standard deviation of feature $f_i$ for class $k$ and $l$; respectively.

Next the features are ranked in a descending order according to their corresponding divergence values and then the correlation between each pair of features is calculated. A threshold of ± 0.8 is set for the correlation values; therefore if a certain correlation value was found to be greater than the specified threshold, the feature with the lower divergence was excluded while the order of the remaining features is preserved. Moreover, independent features (i.e. correlation equal zero) are excluded as well. Hence, only the features that maximise the separability (i.e. with highest divergence) between the different subtypes are kept. For instance, fusing texture features extracted via CM and FD methods from the meningioma subtypes in Fig. 2 – listed in Table II – resulted in up to 37 different features, where each feature vector was labelled with a different index (e.g. index 1 refers to CM contrast feature acquired with 0° angle, index 2 for CM contrast 45° … and so on). All CM features had their $\delta$ set to one, and the FD features are derived after generating an FD image for each subtype as discussed in the fractals section. These features are then ranked according to their divergence values as shown in Table III. Finally a 37×37 correlation matrix is generated and the optimum features are selected by discarding highly correlated (or redundant) features that exceed the set threshold. The redundant features have very limited contribution towards adding information and are merely considered as added noise to the classifier. The features in bold in Table III are the optimised ones, which are CM MP135°, CM IDM45°, FD Lacunarity and CM ENG45°.

This procedure achieved an 89% reduction in the dimensionality of the CM & FD combined feature vector. Additionally, examining the optimised features, it also shows that extracting CM features with different directions – 45° and 135° the best for this case – rather than with a specific direction; or deriving statistics from a generated FD image, such as lacunarity, rather than simply using the mean FD value alone, creates a larger bank of features which broadens the options for selecting the best features that would give a more effective tissue representation (i.e. assisting the classifier by providing high quality features).

At the end of this stage, we will have five texture measures indicated by $X_k$ with their selected optimum features $f_{ki}$, $X_C$ is the fusion of $X_k$ in different combinations.

$$X_C = [X_1(f_{11}, f_{12} \dots f_{1N}) | X_2(f_{21}, f_{22} \dots f_{2N}) | \dots X_k(f_{k1}, f_{k2} \dots f_{ki})] \qquad (11)$$

**Table II Labelled grey level co-occurrence matrix and fractal dimension texture features**

| Index | Texture features |
|---|---|
| 1-4 | CM contrast (0°,45°,90°&135°) |
| 5-8 | CM correlation (0°,45°,90°&135°) |
| 9-12 | CM energy (0°,45°,90°&135°) |
| 13-16 | CM entropy (0°,45°,90°&135°) |
| 17-20 | CM homogeneity (0°,45°,90°&135°) |
| 21-24 | CM dissimilarity (0°,45°,90°&135°) |
| 25-28 | CM inverse difference moment (0°,45°,90°&135°) |
| 29-32 | CM max probability (0°,45°,90°&135°) |
| 33-37 | FD (mean, variance, skewness, kurtosis & lacunarity ) |

**Table III Sorted texture features of Table II in descending order according to corresponding divergence**

| Sorted texture features divergence | | | |
|---|---|---|---|
| Index | value | Index | value |
| **32** | **4.8247** | 4 | 0.7932 |
| 16 | 4.6746 | 36 | 0.3330 |
| 27 | 4.0484 | 29 | 0.2445 |
| 11 | 3.9767 | 13 | 0.2268 |
| 8 | 3.8932 | 5 | 0.1803 |
| 3 | 3.5729 | 21 | 0.1627 |
| 24 | 3.3887 | **26** | **0.0819** |
| 19 | 3.3404 | **37** | **0.0797** |
| 30 | 2.1359 | 34 | 0.0607 |
| 14 | 1.9339 | **10** | **0.0446** |
| 22 | 1.7574 | 2 | 0.0187 |
| 6 | 1.5981 | 35 | 0.0131 |
| 17 | 1.4529 | 18 | 0.0103 |
| 25 | 1.3185 | 33 | 0.0099 |
| 9 | 1.2477 | 31 | 0.0001 |
| 1 | 1.0658 | 15 | 0 |
| 20 | 0.8399 | 7 | 0 |
| 28 | 0.8384 | 23 | 0 |
| 12 | 0.8299 | | |

2.6 Pattern classification technique

The naïve Bayesian classifier is a simple and fast probabilistic classifier which assumes attributes are independent. Yet it is a robust method with on average has a good classification accuracy performance, and even with possible presence of dependent attributes [34], which suits histopathological texture where not all features are conditionally independent given the class. From Bayes' theorem,

$$P_i(C_i|X) = \frac{P(X|C_i)P(C_i)}{P(X)} \qquad (12)$$

Given an image sample $X$ which represent the extracted texture features vector $(f_1, f_2, f_3 \ldots f_n)$ having a probability density function (PDF) $P(X|C_i)$, we tend to maximise the posterior probability $P(C_i|X)$ (i.e., assign sample $X$ to the class $C_i$ that yields the highest probability value).

Here, $P(C_i|X)$ is the probability of assigning class $i$ given feature vector $X$; and $P(X|C_i)$ is the probability of $X$ existing in the given class $i$; $P(C_i)$ is the probability that class $i$ occurs in all the data-set; $P(X)$ is the probability of occurrence of feature vector $X$ in the data-set.

$P(C_i)$ and $P(X)$ can be ignored since we assume that all are equally probable for all samples. This yields the maximum of $P(C_i|X)$ is equal to the maximum of $P(X|C_i)$ and can be estimated using maximum likelihood after assuming a Gaussian PDF [24] as follows:

$$P(X|C_i) = \frac{1}{(2\pi)^{n/2}|\Sigma_i|^{1/2}} exp\left[-\frac{1}{2}(X-\mu_i)^T \Sigma_i^{-1}(X-\mu_i)\right] \quad (13)$$

where $\Sigma_i$ and $\mu_i$ are the covariance matrix and mean vector of feature vector $X$ of class $C_i$; $|\Sigma_i|$ and $\Sigma_i^{-1}$ are the determinant and inverse of the covariance matrix; and $(X-\mu_i)^T$ is the transpose of $(X-\mu_i)$.

The 320 samples which refer to 20 patients were equally divided into four diagnostic groups (i.e. the four meningioma subtypes), each group consists of 80 samples extracted from five different patients (16 each) diagnosed with the same meningioma tumour subtype. Since the number of image subsets is not small, a holdout validation approach was used for validation of classification, by randomly selecting four patients from each group for training and the remaining for testing.

## 3. Experimental Results

### 3.1 Colour channel selection

The results of the Bhattacharya distance which specifies the segmentation quality for the three colour channels of the four meningioma subtypes are shown in Table IV, with the smallest (i.e. most separable) values in bold. All of the meningioma subtypes except the fibroblastic had a better segmentation quality using the blue colour channel.

To investigate the suitability of the blue colour channel for automated classification of the histopathological images used in this work, the meningioma images were further represented in another three colour spaces to determine which channel would best characterise the tissue texture. Other commonly used HSV, CIE-$L^*a^*b^*$ and YCbCr colour spaces in the image processing literature were applied in addition to the RGB, and each of the channels were used for classification after applying the morphological gradient. The colour channel selection was reflected in the classification accuracies for the texture measures. Table V shows the RLM RGB colour channel classification accuracy for the four meningioma subtypes, with the blue channel achieving the highest on overall. Similar results were obtained for the rest of the used texture measures, for succinctness they are not presented here.

Table IV Assessing classification quality for each colour channel

| Meningioma type | Red | Green | Blue |
|---|---|---|---|
| Fibroblastic | **0.0010** | 0.0082 | 0.0161 |
| Meningothelial | 0.0066 | 0.0032 | **0.0020** |
| Psammomatous | 0.1261 | 0.2045 | **0.1250** |
| Transitional | 0.0016 | 0.0081 | **0.0011** |

**Table V The RGB colour channels classification accuracies for the RLM texture measure**

|  |  | Meningioma type | | | | |
|---|---|---|---|---|---|---|
|  |  | Fibroblastic | Meningothelial | Psammomatous | Transitional | Overall accuracy |
| Various colour space channels | R | 80.00% | 95.00% | 80.00% | 70.00% | 81.25% |
|  | G | 95.00% | 80.00% | 90.00% | 50.00% | 78.75% |
|  | B | 90.00% | 75.00% | 85.00% | 85.00% | **83.75%** |
|  | H | 100% | 55.00% | 90.00% | 65.00% | 77.50% |
|  | S | 70.00% | 75.00% | 60.00% | 85.00% | 72.50% |
|  | V | 90.00% | 85.00% | 75.00% | 60.00% | 77.50% |
|  | $L^*$ | 80.00% | 75.00% | 80.00% | 50.00% | 71.25% |
|  | $a^*$ | 50.00% | 70.00% | 75.00% | 80.00% | 68.75% |
|  | $b^*$ | 90.00% | 40.00% | 75.00% | 70.00% | 68.75% |
|  | Y | 80.00% | 70.00% | 90.00% | 60.00% | 75.00% |
|  | $C_b$ | 95.00% | 55.00% | 75.00% | 75.00% | 75.00% |
|  | $C_r$ | 65.00% | 60.00% | 80.00% | 95.00% | 75.00% |

*3.2 Individual and combined classification accuracies*

Testing classification accuracies for each of the individual textures and in different combinations are as shown in Table VI and VII; respectively. These results represent the morphological gradient images of the blue colour component meningioma images for all four types.

When selectively combining certain texture features, the classification accuracy would increase above the highest achieved if an individual texture features method was used alone. For example, Table VI shows the overall classification accuracies if the extracted texture features would be used individually (i.e. without combining them with each other). The RLM texture feature achieved the highest overall accuracy by 83.75%. Yet, when fusing the texture features with each other, and in all possible combinations, some combination improved the overall accuracy up to 92.50% as in the GMRF & RLM paired features (as shown in Table VII). By taking the RLM classification accuracy from Table VI and setting it as a threshold – as it achieved the highest in case if each texture feature was used individually – then we can see that the first four rows in Table VII for the combined texture feature improved the accuracy. To investigate the significance of the results, a Wilcoxon signed-rank test – a nonparametric equivalent to the paired t-test – was applied to determine the significance between the texture measure combinations that improved the overall accuracy and the individual approaches. The test shows there is statistically significant difference on a significance level of $p < 0.05$.

The optimum texture features for each of the texture features combinations that improved the classification accuracy are ranked according to their divergence power and listed in Table VIII. For completeness, the confusion matrix for the best combination (RLM & GMRF) is given in Table IX.

**Table VI Individual texture features testing classification accuracy The blue colour component of Meningioma images**

| Meningioma type | FD | RLM | CM | GMRF |
|---|---|---|---|---|
| Fibroblastic | 35.00% | 90.00% | 75.00% | 70.00% |
| Meningothelial | 50.00% | 75.00% | 75.00% | 90.00% |
| Psammomatous | 75.00% | 85.00% | 90.00% | 70.00% |
| Transitional | 65.00% | 85.00% | 80.00% | 80.00% |
| **Overall accuracy** | **56.25%** | **83.75%** | **80.00%** | **77.50%** |

**Table VII Classification accuracy of extracted texture features in different combinations ranked in descending order**

| Texture features | Fibroblastic | Meningothelial | Psammomatous | Transitional | Overall accuracy |
|---|---|---|---|---|---|
| **GMRF&RLM** | **90.00%** | **95.00%** | **90.00%** | **95.00%** | **92.50%** |
| **GMRF&FD&RLM** | **80.00%** | **80.00%** | **90.00%** | **100.00%** | **87.50%** |
| **RLM&FD** | **80.00%** | **80.00%** | **90.00%** | **95.00%** | **86.25%** |
| **GMRF&FD&CM** | **90.00%** | **75.00%** | **90.00%** | **85.00%** | **85.00%** |
| GMRF&CM | 80.00% | 75.00% | 85.00% | 85.00% | 81.25% |
| GMRF&RLM&CM | 70.00% | 90.00% | 85.00% | 80.00% | 81.25% |
| RLM&CM | 60.00% | 95.00% | 85.00% | 80.00% | 80.00% |
| RLM&FD&CM | 80.00% | 90.00% | 80.00% | 80.00% | 80.00% |
| GMRF&RLM&FD&CM | 80.00% | 70.00% | 90.00% | 80.00% | 80.00% |
| FD&CM | 80.00% | 75.00% | 90.00% | 70.00% | 78.75% |
| GMRF&FD | 70.00% | 70.00% | 90.00% | 70.00% | 75.50% |

**Table VIII Optimum features for the top four texture combination features in Table VI which improved the classification accuracy beyond the set threshold**

| Texture features | Index | Divergence | Optimum features |
|---|---|---|---|
| GMRF&RLM (51)[†] | 5 | 1.8543 | RLM LRE0° |
| | 49 | 0.0011 | GMRF sxy;5 |
| | 51 | 0.0008 | GMRF $\sigma$ |
| | 36 | 0.0001 | RLM SRHGLE135° |
| GMRF&FD&RLM (56) | 5 | 1.8543 | RLM LRE0° |
| | 49 | 0.0011 | GMRF sxy;5 |
| | 51 | 0.0008 | GMRF $\sigma$ |
| | 36 | 0.0001 | RLM SRHGLE135° |
| | 52 | 0.0000 | FD mean |
| RLM&FD (49) | 5 | 1.8543 | RLM LRE0° |
| | 26 | 0.0001 | RLM HGLRE45° |
| | 36 | 0.0001 | RLM SRHGLE135° |
| | 45 | 0.0000 | FD mean |
| | 29 | 0.0000 | RLM SRGLE0° |
| GMRF&FD&CM (44) | 32 | 1.7274 | CM MP135° |
| | 37 | 1.0535 | GMRF sxy;5 |
| | 39 | 0.8265 | GMRF $\sigma$ |
| | 43 | 0.4548 | FD kurtosis |
| | 26 | 0.0722 | CM IDM45° |

**Table IX Four class meningioma classification confusion matrix for the combined GMRF and RLM texture measures**

| Meningioma type | | Classification | | | |
|---|---|---|---|---|---|
| | | Fibroblastic | Meningothelial | Psammomatous | Transitional |
| True class | Fibroblastic | 90% | 5% | 0% | 0% |
| | Meningothelial | 0% | 95% | 0% | 5% |
| | Psammomatous | 0% | 0% | 90% | 0% |
| | Transitional | 10% | 0% | 10% | 95% |

---

[†] Number between brackets indicates the total number of combined texture features before optimum feature selection

*3.3 Morphological processing*

A comparison between the effect of morphological and non-morphological processing on the image classification accuracy is shown in Fig. 7. All MP texture measures except FD witnessed an increase in the overall classification accuracy. We chose to go forward with the morphological gradient option as the FD gave the least classification.

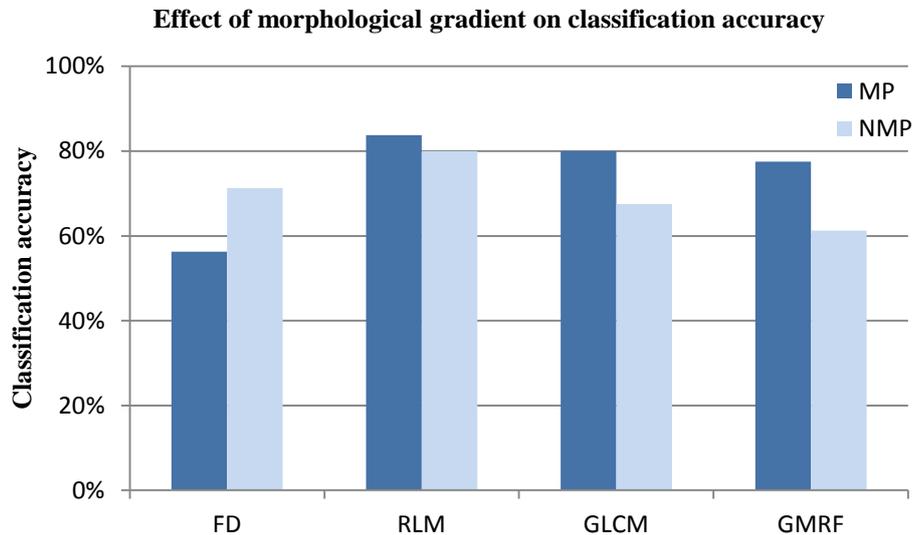

Fig. 7 Individual texture measures' classification accuracy for morphologically processed (MP) and non-morphologically processed (NMP) of the four class meningioma images.

## 4. Texture Measures Behaviour Analysis

*4.1 Relevance to histopathological texture*

To explain why certain combinations of texture features used in this work tend to work better, we applied the best three texture features that improved the overall accuracy to a set of 15 different generated images having jelly-bean shapes resembling in analogy the shape of the cell nuclei in the meningioma images (see Fig. 8). These pseudo-cell nuclei images start with a specific number of similar shapes in the first image, and then increase gradually by an amount equivalent to the number of shapes in the first image, until reaching the last image in the set. By this, we intend to see how the performance of each texture measure is affected as the frequency of the examined structure changes. Then the morphological gradient is computed for each of these images as was done with the real meningioma images.

In reality, the different images of a certain type of meningioma do not necessarily have an identical structure or the same number of cell nuclei. Thus, it is interesting to know how the applied texture measures cope with this situation and how their performance is affected. Hence one can better understand why certain combinations might work better.

The variation is assessed by measuring the mean ($\mu$) and standard deviation ($\sigma$) of the extracted texture features and then representing them by the ratio $\sigma/\mu$ which would reflect the susceptibility of the examined texture measure to the increase and decrease of the frequency (i.e. denseness) of the examined structure. Clausi et al studied the effect of Gaussian additive noise on Gabor filters and CM texture features together, they showed that CM is less susceptible to noise as compared to Gabor filter [35]. Yet we use this ratio first to investigate the structure denseness impact on the used texture measures, and then in the next subsection the noise effect is presented. It should be noted that, as explained in the feature extraction section, the RLM is represented by 11 features, GMRF by 7 and the FD by 5. So the ratio $\sigma/\mu$ would represent the joint effect of all extracted features relevant to each texture measure.

The best feature combinations in Table VII that improved the overall accuracy are normalised and plotted. Fig. 9 represents the value of the ratio $\sigma/\mu$ for each of the 15 generated images, which is interpreted as the lower the ratio the less susceptible the texture measure. It is noticed that the GMRF extracted features are nearly uniform throughout the image set, with a vey slight increase in the variation as frequency increases. On the contrary, the RLM and FD show more susceptibility in low frequencies and are less susceptible in the high frequency as compared to GMRF. It is also shown that the RLM performs better in the low and high frequencies as compared to FD, while the FD is more stable in the mid range frequencies.

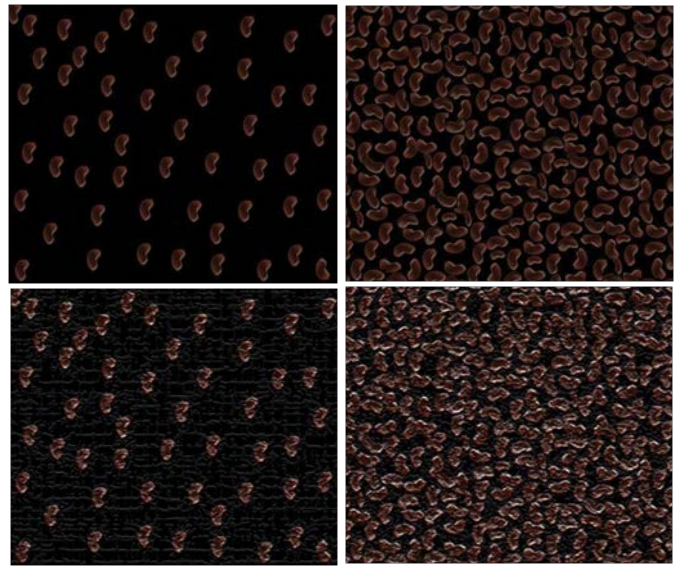

**Fig. 8 The first row is the initial and final images in the 15 generated pseudo-cell nuclei images to test texture measures susceptibility to variation in shape frequency, the second row is the corresponding distorted images.**

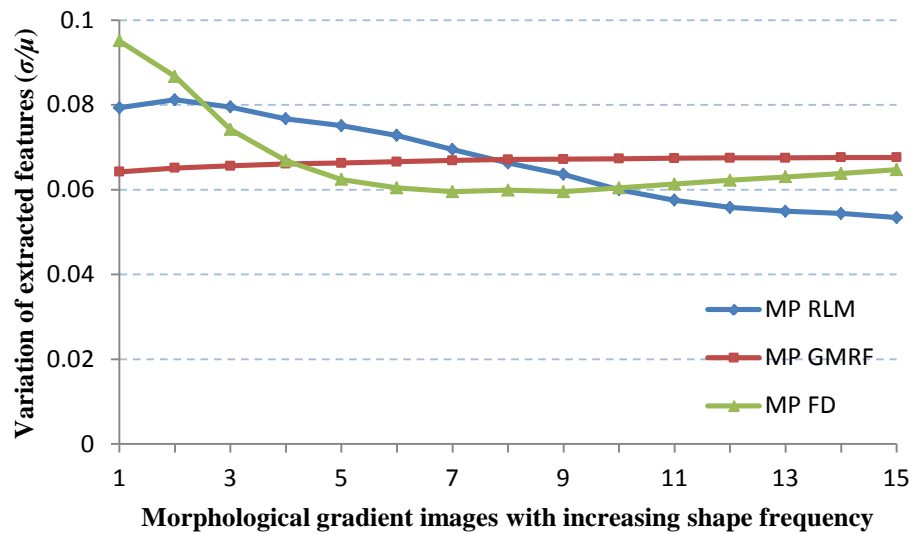

**Fig. 9 Susceptibility of RLM, GMRF and FD texture measures to 15 morphologically processed pseudo-cell nuclei images with increasing shape frequency.**

*4.2 Simulation of noise impact on extracted features*

To study the impact of noise on the meningioma images, distortion having an effect resembling fine cracks or craquelures which appear on old paintings was applied to 15 pseudo-cell nuclei images (see Fig. 8). As the most probable noise to affect the histopathological images in the process of preparation is the cracks in the biopsy sample which is most obvious as the white regions in the psammomatous meningioma images in Fig. 2. A one dimensional horizontal cross section in a pseudo-cell nuclei image before and after noise distortion is shown in Fig. 10.

Although the applied morphological gradient eliminates the background – including the white cracks in the image sample (see upper right corner of Fig. 5) - to extract the general cell nuclei structure, these cracks can still alter the general shape that a certain type of meningioma cell nuclei should take.

Analysing the susceptibility of the texture measures as shown in Fig. 11, the GMRF was the least affected by the added noise as it gave nearly a uniform response throughout all images. The RLM and FD behaved oppositely to each other in response to noise in a monotonically decreasing/increasing fashion, respectively. In a way, RLM is less susceptible to noise in high shape frequency of occurrence; vice versa for FD. Therefore the response of each texture measure somehow depends on the structure that the noise affects. For example if the noise occurs in dense image structures, the RLM measure could produce a more reliable estimate as compared to the other measures. We can also see that roughly similar noise susceptance is produced in between pseudo-cell nuclei image 7 and 8 in Fig. 11 for all three texture measures.

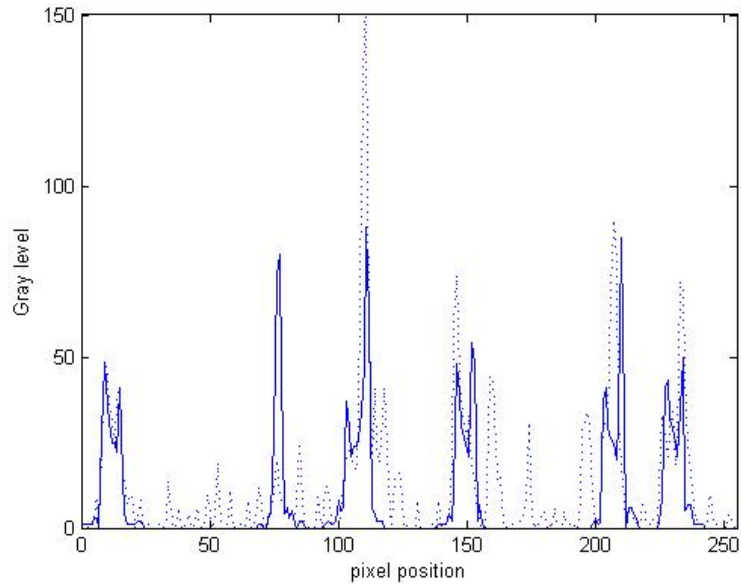

**Fig. 10** One dimensional horizontal grey-level profile along the first pseudo-cell nuclei image, the dotted line indicates the profile of the effect of added craquelures distortion.

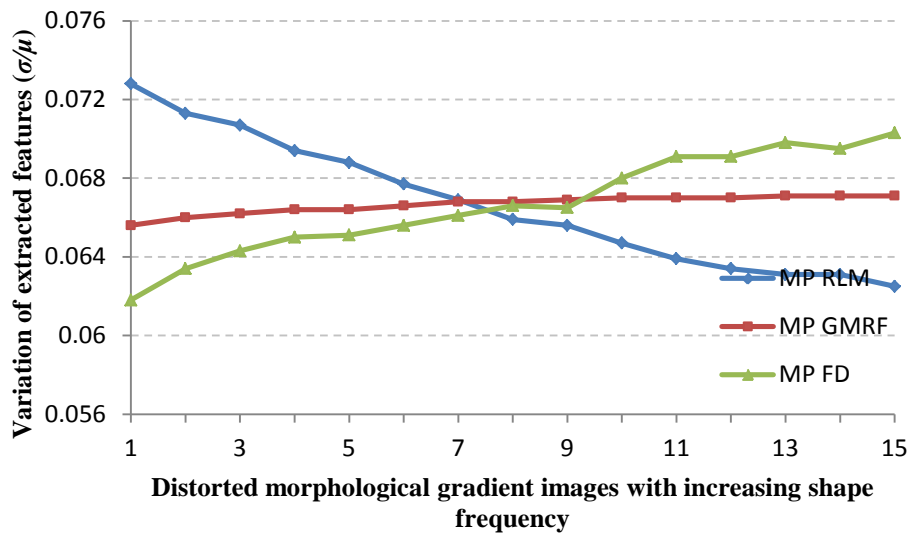

**Fig. 11** Effect of noise distortion on texture measures applied to images in Fig2.

## 5. Discussion

The main objective was to improve meningioma histopathological image classification accuracy to overcome inter-and- intra-observer variations in human reported diagnosis. The technique exploits the physiological structure of the cell nuclei with four different texture measures, and attempts to find the best combination that maximises the difference between the meningioma subtypes.

It was shown that the choice of colour channel can affect the classification accuracy, in terms of which better defines the borders of the region of interest – the cell nuclei in our case – from the background. The quality of segmentation performed on a sample image from each subtype favoured the blue colour channel for three of the four subtypes. This is due to the dyes used in staining the meningioma biopsies which gave the cell nuclei a purple colour and the background (i.e. cytoplasm) a pink colour, where the better segmentation performance can be interpreted as the dominance of blue component in the purple colour which consists mainly of mixtures of blue and red. Usually the fibroblastic subtype is harder to differentiate from other subtypes [36], furthermore, the relatively small size and the faint colour of some of the fibroblastic cell-nuclei as compared to the other subtypes contributed towards giving the red colour channel a better separability. Yet, the subtypes overall classification results showed that the blue colour channel was the best for all subtypes.

Also it was shown that combining more than one texture measure instead of using just one might improve the overall accuracy. Different texture measure tends to extract different features each capturing alternative characteristics of the examined structure. The two model and statistical-based texture measures (GMRF and RLM) improved the overall accuracy up to 92.50% with none of the classified meningioma subtypes achieving below 90.00%. As indicated in the confusion matrix, we see that all misclassified subtypes are related only to a single other subtype (e.g. fibroblastic and psammomatous subtypes were always misclassified as transitional). The misclassification occurs mainly due to non-homogeneity of the cell nuclei structure for the prepared biopsies. Another reason is in the subdivision of each of the 1024x1024 pixel images to four quadrants (i.e. subsets), the shape of the structure in some of the four 512x512 pixel quadrants and with the possible presence of some biopsy preparation cracks in that quadrant, might not be sufficient to capture the original subtype cell-nuclei shape, and hence be more likely to be misclassified.

Nevertheless, using a combination of multiple texture measure does not necessarily guarantee a better accuracy, even with the removal of highly correlated features. All four texture measures combined – appearing in the ninth row of Table VII – gave an 80%, degrading the overall classification accuracy below the 83.75% set threshold. This implies fewer (paired) texture measures could best characterise the examined texture and produce far better classification results in a shorter CPU processing time.

Moreover, classification results suggest that taking the morphological gradient for the histopathological images would serve most texture measures' capability to capture tissue characteristics; yet, the stability of the texture measures' response varies depending on the examined structure shape denseness. By studying the variation of the texture measure features as the number of cell-nuclei increases, the GMRF was nearly uniform, while the RLM and FD performed better in the high frequencies. That is, in the GMRF and RLM combined, the RLM is less affected (i.e. the variation of the normalised features is less as compared to GMRF features) if the number of cell-nuclei increase suddenly in on of the examined samples above the expected average, which will assist in classifying it correctly. In a way, they compensate for each others' weaknesses.

Varying amounts and types of noise is inevitable in medical imaging which will have some effect on used texture measures [37]. Fine cracks in the tissue biopsies are a major source of noise that can affect histopathological images. The texture measures' response to additive texture distortion noise whilst varying cell-nuclei shape densities was studied. The GMRF was the least affected, yet the RLM and FD performed better in high and low shape frequency; respectively.

A limitation of the proposed meningioma classification technique is that segmentation separability assessment is required in order to select the optimum colour channel, yet we need this process only once (i.e. before training).

## 6. Conclusion

A technique for histopathological meningioma tumour classification based on texture measures combination, which aims to overcome intra and inter-observer variability, has been proposed in this study. The morphological gradient of the RGB colour channel that best discriminates the cell-nuclei from the cytoplasm background is selected, and then feature extraction is performed by four statistical and model-based texture measures for discrimination using a Bayesian classifier. The pre-processing phase represented by the appropriate colour channel selection and morphological processing proved to be necessary for increasing texture feature separability, and hence can improve classification accuracy.

It can be concluded that certain selected texture measures play a complementary role to each other in the process of quantitative texture characterisation. In other words, a certain texture measure can represent a pattern better than another depending on the region of interest frequency of occurrence and noise in the examined structure. This also applies to certain combinations which might outperform other texture measure fusions. However, combining more than two texture measures would not necessarily give a better accuracy even with the removal of highly correlated features. This will increase feature complexity, hence having a negative effect on the classifier's performance. It was found that the combination of the GMRF & RLM texture measures are the best for characterizing meningioma subtypes of grade I, these two measures outperformed other measures in the study individually and combined. Furthermore, it would be interesting to test the compatibility of the suggested meningioma classification approach to discern in-between subtypes and/or grades of other similar histopathological diseases.

## Acknowledgments

The author would like to thank Prof. Volkmar Hans from the Institute of neuropathology in Bielefeld, Germany, for providing the histopathological data-set used in this work.